\documentclass[final,5p,times,twocolumn,authoryear]{elsarticle}

\usepackage{amssymb}
\usepackage{bm}
\usepackage{color}

\journal{arXiv}

\begin{document}

\begin{frontmatter}



\title{Activation Functions for ``A Feedforward Unitary Equivariant Neural Network"}


\author[label1]{Pui-Wai Ma}
\ead{Leo.Ma@eng.ox.ac.uk}
\affiliation[label1]{organization={Department of Engineering Science,University of Oxford},
            addressline={Parks Road}, 
            city={Oxford},
            postcode={OX1 3PJ}, 
            country={United Kingdom}}


\begin{abstract}
In our previous work [\cite{MA2023154}], we presented a feedforward unitary equivariant neural network. We proposed three distinct activation functions tailored for this network: a softsign function with a small residue, an identity function, and a Leaky ReLU function. While these functions demonstrated the desired equivariance properties, they limited the neural network's architecture. This short paper generalises these activation functions to a single functional form. This functional form represents a broad class of functions, maintains unitary equivariance, and offers greater flexibility for the design of equivariant neural networks.
\end{abstract}




\end{frontmatter}



Activation functions play a crucial role in the design and performance of neural networks. They introduce non-linearity into the model, enabling the network to learn complex patterns. 

We proposed a feedforward unitary equivariant neural network [\cite{MA2023154}]. We also proposed three activation functions specifically designed for such a neural network: a modified softsign function, an identity function, and a Leaky ReLU function. These functions were chosen to preserve the unitary equivariance of the network, a property that ensures the network's output remains consistent under unitary transformations of the input.

The first one is a softsign function with a small residue, given by 
\begin{equation}
    \bm{\sigma}(\mathbf{u}) = \frac{\mathbf{u}}{1+||\mathbf{u}||} + a\mathbf{u},
\end{equation}
where $a\in\mathbb{R}^{\geq 0}$ is a (small) scalar constant, and $\mathbf{u}\in \mathbb{C}^n$. $||\mathbf{u}||\in\mathbb{R}$ is the norm of $\mathbf{u}$.

The second one is the identity function, defined as
\begin{equation}
    \bm{\sigma}(\mathbf{u})=\mathbf{u}.
\end{equation}

The third one is a Leaky ReLU function, defined as 
\begin{equation}
     \bm{\sigma}(\mathbf{u})=
     \left\{
     \begin{array}{cc}
     \mathbf{u} &  \texttt{if } ||\mathbf{u}|| \geq c,\\
     k\mathbf{u} & \textit{otherwise},
     \end{array}
     \right.
\end{equation}
where $1 > k \in \mathbb{R}^{\geq 0}$ and $c\in \mathbb{R}^{\geq 0}$ are positive scalar constants.  

However, if only three activation functions are allowed in the design of a neural network, this is overly restrictive. 

By inspecting the structure of these three activation functions, we can deduce a generalized functional form for the activation function, given by
\begin{equation}
    \bm{\sigma}(\mathbf{u}) = f(x)\mathbf{u},
\end{equation} 
where $f:\mathbb{R}\rightarrow\mathbb{R}$ is a scalar function, and $x:\mathbb{C}^n \rightarrow \mathbb{R}$, such that
\begin{equation}
    x=x(\mathbf{u})=x(\mathcal{U}\mathbf{u}),
\end{equation}
for any unitary operator $\mathcal{U}$. A choice of $x$ could be 
\begin{equation}
    x = ||\mathbf{u}|| - \kappa,
\end{equation}
and $\kappa\in\mathbb{R}$ is a constant. $f$ can be chosen as any commonly used scalar activation function, such as the sigmoid, hyperbolic tangent, Leaky ReLU, etc. Since $||\mathbf{u}||>0$, the introduction of $\kappa$ allows $x$ to have negative values. $\kappa$ can be obtained as a parameter during model training and may vary for different nodes. 

We can immediately verify that the activation function satisfies the following:
\begin{equation}
    \bm{\sigma}(\mathcal{U}\mathbf{u})=f(x)\mathcal{U}\mathbf{u}=\mathcal{U}f(x)\mathbf{u}=\mathcal{U}\bm{\sigma}(\mathbf{u}),
\end{equation}
which is a requirement of our feedforward unitary equivariant neural network. 

This generalised form not only preserves unitary equivariance but also allows for a broader range of functions, potentially enhancing the network's capacity to learn and represent complex data. This theoretical advancement opens up new possibilities for the design and implementation of neural networks, particularly in applications where unitary transformations are used.


\section*{Acknowledgments}
The author would like to acknowledge the visiting fellow status granted by the Department of Engineering Science, University of Oxford. 

\bibliographystyle{elsarticle-harv} 


\end{document}